\documentclass[manuscript,screen]{acmart}

\AtBeginDocument{%
  }

\setcopyright{acmlicensed}
\copyrightyear{2024}
\acmYear{2024}
\acmDOI{XXXXXXX.XXXXXXX}

\acmISBN{978-1-4503-XXXX-X/18/06}

\usepackage{enumitem}
\usepackage{graphicx}%
\usepackage{multirow}%
\usepackage{mathrsfs}%
\usepackage[title]{appendix}%
\usepackage{xcolor}%
\usepackage{textcomp}%
\usepackage{manyfoot}%
\usepackage{booktabs}%
\usepackage{algorithm}%
\usepackage{algorithmicx}%
\usepackage{algpseudocode}%
\usepackage{listings}%

\begin{document}

\title{DPERC: Direct Parameter Estimation for Mixed Data}


\author{Tuan L. Vo}
\email{tuanlvo293@gmail.com}
\affiliation{%
  \institution{University of Science, Vietnam National University Ho Chi Minh City}
  \city{Ho Chi Minh}
  \country{Vietnam}
}

\author{Quan Huu Do}
\affiliation{%
  \institution{University of Science, Vietnam National University Ho Chi Minh City}
  \city{Ho Chi Minh}
  \country{Vietnam}
}

\author{Uyen Dang}
\affiliation{%
  \institution{University of Science, Vietnam National University Ho Chi Minh City}
  \city{Ho Chi Minh}
  \country{Vietnam}
}

\author{Thu Nguyen}
\affiliation{%
 \institution{Simula Metropolitan}
 \city{Oslo}
 \country{Norway}}

\author{Pål Halvorsen}
\affiliation{%
 \institution{Simula Metropolitan}
 \city{Oslo}
 \country{Norway}}

\author{Michael A. Riegler}
\affiliation{%
 \institution{Simula Metropolitan}
 \city{Oslo}
 \country{Norway}}

\author{Binh T. Nguyen}
\affiliation{%
  \institution{University of Science, Vietnam National University Ho Chi Minh City}
  \city{Ho Chi Minh}
  \country{Vietnam}
}
\authornote{Corresponding Author}
\email{ ngtbinh@hcmus.edu.vn}


\begin{abstract}
 The covariance matrix is a foundation in numerous statistical and machine-learning applications such as Principle Component Analysis, Correlation Heatmap, etc. However, missing values within datasets present a formidable obstacle to accurately estimating this matrix. While imputation methods offer one avenue for addressing this challenge, they often entail a trade-off between computational efficiency and estimation accuracy. Consequently, attention has shifted towards direct parameter estimation, given its precision and reduced computational burden. In this paper, we propose Direct Parameter Estimation for Randomly Missing Data with Categorical Features (DPERC), an efficient approach for direct parameter estimation tailored to mixed data that contains missing values within continuous features. Our method is motivated by leveraging information from categorical features, which can significantly enhance covariance matrix estimation for continuous features. Our approach effectively harnesses the information embedded within mixed data structures. Through comprehensive evaluations of diverse datasets, we demonstrate the competitive performance of DPERC compared to various contemporary techniques. In addition, we also show by experiments that DPERC is a valuable tool for visualizing the correlation heatmap.
\end{abstract}



\keywords{missing data, maximum likelihood estimate, parameter estimation}


\maketitle

\section{Introduction}\label{sec:introduction}

Data are collected in various ways, such as paper-based and online surveys, interviews, and sensors. So, the data is usually incomplete. In fact, missing data happens for many reasons. For example, temperature and humidity data are often acquired through different sensors. Because of a variety of factors, such as human error, misunderstanding, and equipment malfunctions, data may frequently experience loss or disturbance. This leads to missing data, which can hinder analysis, modeling, and the explainability of the built model. A critical task in case of missing data is estimating the covariance matrix, which provides essential information about the relationships between different variables in a dataset, and this is usually reflected in the correlation plot~\cite{pham2024correlation}. Also, various applications, such as hypothesis testing, modeling uncertainty with confidence intervals, parameter estimation, and principal component analysis, also rely on the estimation of the covariance matrix. Even minor enhancements in covariance matrix estimation can substantially impact the quality and reliability of statistical analyses and machine learning models. In finance, for instance, a precise covariance matrix is indispensable for tasks such as risk management~\cite{deshmukh2019improved} and portfolio optimization~\cite{math10224282, RePEc:mse:cesdoc:19022}. Slight improvements in covariance matrix estimation result in better precision matrix estimation~\cite{vo2024effects} and more accurate predictions of asset behaviors and correlations, thereby enhancing the performance of financial models. This improvement translates into more effective risk assessment and better-optimized investment strategies, ultimately leading to higher returns and reduced exposure to risk. In fields like signal processing, improved covariance matrix estimation can enhance techniques for noise reduction, facilitating clearer extraction of signals~\cite{de_Santi_2022}.

For handling missing data, it is proven that the two-step procedures (imputation - parameter estimation) can be computationally expensive compared to direct parameter estimation~\cite{nguyen2021epem,NGUYEN2022108082}. Specifically, the DPER algorithm~\cite{NGUYEN2022108082} performs well in estimating the covariance matrix when data only come from continuous features. However, this algorithm only relies on continuous features for estimation and does not exploit the categorical features under mixed data. We, however, hypothesize that categorical can provide important information that aids the quality of the estimation for continuous features.
Therefore, we present the DPERC algorithm (Direct Parameter Estimation for Randomly missing data with Categorical features) as an endeavor to enhance DPER for a single class dataset. This is done by selecting a categorical to maximize the log-likelihood, treating it as an artificial label, and applying the DPER algorithm \cite{NGUYEN2022108082}.

The contribution of our paper can be summarized as follows:
\begin{enumerate}

    \item We propose DPERC, an improvement over DPER for mixed-type data for two following situations: (i) the data comes from a single class; (ii) the data comes from multiple classes without the assumption of equal covariance matrices,
    \item We provide the theoretical ground for the proposed technique,
    \item  We illustrate that DPERC achieves better performance compared to method comparison and shows an improvement compared to DPER in covariance matrix estimation tasks,
    \item We demonstrate with experiments that DPERC and DPER can provide correlation heatmaps better than various imputation techniques.
\end{enumerate}
The rest of this article is organized as follows.  
Section~\ref{sec:related_work} overviews relevant recent work on incomplete data. Next, we provide background on the direct parameter estimation for randomly missing data (DPER) in Section~\ref{sec:preli}. In Section~\ref{sec:DPERC single}, we present the main ideas and theoretical grounds for the algorithm DPERC and then present the algorithm DPERC for single-class and multiple-class mixed data. After that, in Section~\ref{sec:experiments}, we illustrate the efficiency of our algorithm via experiments and analyze the results. Finally, the paper ends with conclusions and our future works in Section~\ref{sec:conclusion}.

\section{Related works}\label{sec:related_work}

Due to the prevalence of missing data, numerous methods have been proposed to address this issue. Typically, there are two major approaches for the parameter estimation probability: (1) utilizing imputation methods and then estimating the covariance matrix from imputed data; (2) directly estimating parameters from missing data, which can be summarized as follows.

The first approach uses imputation methods to fill in missing values before estimating the covariance matrix. In order to commence with single imputation methods, these approaches involve techniques that take advantage of the temporal correlation between data, thus imputing missing values using the available statistical values such as mean, mode, and median~\cite{donders2006gentle}. Other methods based on statistics learning, such as Hot-deck~\cite{andridge2010review}, impute missing values within a data matrix by using available values from the same matrix; Multiple imputation~\cite{rubin2018multiple}, addresses missing data by creating reasonable estimates based on the patterns and connections among the available variables in the dataset. Various machine-learning techniques have been utilized to fill in missing data. For example, k-nearest neighbor imputation (KNN~\cite{beretta2016nearest},\cite{sanjar2020missing}) utilizes k-nearest neighbors (KNN) algorithm to estimate missing based on the characteristics of similar neighboring data points. As another example, multiple Imputation by Chained Equations (MICE~\cite{buuren2010mice}) iteratively imputes missing values using regression models based on other variables in the dataset. Next, Bayesian network imputation~\cite{hruschka2007bayesian} leverages probabilistic relationships between variables, while methods such as multiple imputation using Deep Denoising Autoencoders~\cite{audigier2016multiple} employ neural networks to learn intricate data patterns for imputation. In addition, multi-layer perceptron (MLP~\cite{silva2011missing}) and self-organization maps (SOM~\cite{vatanen2015self}) also utilize a neural network architecture to predict missing values in a dataset based on the relationships between observed variables. Another group of techniques includes methods based on matrix completion, where matrix factorization is used to exploit the inherent low-rank structure of the data and thereby infer the missing entries. Typical techniques in this category include Polynomial Matrix Completion~\cite{fan2020polynomial}, SOFT-IMPUTE\cite{mazumder2010spectral}, and Nuclear Norm Minimization~\cite{candes2009exact}.

To handle data that contains both categorical and continuous features, we can first mention the technique based on Random Forest like missForest~\cite{stekhoven2012missforest}, which operates by averaging predictions from numerous unpruned classification or regression trees. This inherent feature of random forests essentially forms a multiple imputation scheme. Modeling data with the Gaussian copula model~\cite{zhao2020missing} has been studied with Bayesian approaches. Another approach is based on a principal component method~\cite{audigier2016principal},  the prediction of the missing values is based on the similarity between individuals and on the relationships between variables. With large datasets that contain many missing values, the imputation method based on multilevel singular value decomposition (SVD~\cite{husson2019imputation}) can be considerable. 

The second approach is estimating the covariance matrix without imputation. This can be done by integrating the information gleaned from observed data with statistical assumptions to formulate estimations of population parameters and/or the statistical mechanism governing missing data~\cite{dong2013principled}. This approach can overcome the drawback of the first one, which can be computationally inefficient. When dealing with high-dimensional data, approximately low-rank covariance matrix estimation~\cite{lounici2014high} can be a convenient option.  The group of methods relies on Maximum Likelihood such as the Expectation-Maximization (EM~\cite{nelwamondo2007missing}) and its improved name Equalization-Maximization (EqM~\cite{stoica2005parameter}); Full Information Maximum Likelihood (FILM~\cite{enders2001relative}), which determines the model parameters by maximizing the likelihood function of the observed data, given those parameters, utilizing all accessible information. Under various assumptions for the data that contains missing values, we have a group of techniques estimating the maximum likelihood estimators for a multivariate normal distribution, consisting of EPEM~\cite{nguyen2021epem}, which can be applied to monotone missing data, PMF\cite{nguyen4260235pmf} when the missing data only occurs in some features, and DPER can be used for randomly missing data. Specifically, EPEM estimates the MLEs for multiple class monotone missing data under the assumption of equal covariance matrices. PMF and DPER are meant for a more general case where missing data randomly occur with/without the assumption of equal covariance matrices. However, when dealing with mixed-type data, DPER and PMF only use the information from continuous features and ignore the categorical features. 

\section{Preliminaries}\label{sec:preli}

In this section, we briefly summarize the DPER algorithm for directly estimating parameters for randomly missing data. The algorithm relies on the following theoretical ground in case data comes from multiple classes under the assumption of equal covariance matrices.

\begin{theorem}\label{bivariatemle_multi} 
Assume that we have a dataset with $G$ classes and the assumption of equal covariance matrices, where each sample from the $g^{th} (1\le g\le G)$ class follows a bivariate normal distribution with mean  $
		\boldsymbol {\mu }^{(g)} = \begin{pmatrix}
			\mu_1^{(g)}\\ \mu_2^{(g)}
		\end{pmatrix}
	$and covariance matrix $
		\boldsymbol {\Sigma } =  \begin{pmatrix}
			\sigma_{11}&\sigma_{12}\\
			\sigma _{21} & \sigma _{22}
		\end{pmatrix},$
	and arrange the data into the following pattern
	\begin{equation*}\label{blform}
		\boldsymbol{x}^{(g)} = \begin{pmatrix}
			{x}_{11}^{(g)} & ... & {x}_{1m_g}^{(g)}&{x}^{(g)}_{1m_g+1}&...&{x}^{(g)}_{1n_g}&*&...&*\\
			{x}^{(g)}_{21}&...&{x}^{(g)}_{2m_g}& *& ...& *& {x}^{(g)}_{2n_g+1}&...&{x}^{(g)}_{2l_g}
		\end{pmatrix}
	\end{equation*}
	So, each column represents an observation, and $x_{ij} \in \mathbb{R}$ is an entry, i.e., each observation has two features.  
	
	Let $L$ be the likelihood of the data and	
	\begin{align*}
		A &= \sum_{g=1}^Gm_g,\\
		s_{11} &= \sum_{g=1}^G\sum_{j=1}^{m_g}(x^{(g)}_{1j}-\widehat{\mu}^{(g)}_1)^2	,\\
		s_{12} &= \sum_{g=1}^G\sum_{j=1}^{m_g}(x^{(g)}_{2j}-\widehat{\mu}^{(g)}_2)(x^{(g)}_{1j}-\widehat{\mu}^{(g)}_1),\\
		s_{22} &= \sum_{g=1}^G\sum_{j=1}^{m_g}(x^{(g)}_{2j}-\widehat{\mu}^{(g)}_2)^2.
	\end{align*}
	
	Then, the resulting estimators obtained by maximizing $L$ w.r.t $\mu^{(g)}_1, \sigma_{11},\mu^{(g)}_2, \sigma_{22}$, and $\sigma_{12}$ are:
	\begin{align*}
		\widehat{\mu}^{(g)}_1 & = \frac{1}{n_g}\sum_ {j=1}^{n_g}x^{(g)}_{1j},\;\;
		\widehat{\mu}^{(g)}_2 = \frac{\sum_ {j=1}^{m_g}x^{(g)}_{2j}+ \sum_ {j=n_g+1}^{l_g}x^{(g)}_{2j}}{m_g+l_g-n_g},	 \\
		\widehat{\sigma}_{11} & = \frac{\sum_{g=1}^G\sum_ {j=1}^{n_g}(x^{(g)}_{1j}-\widehat{\mu}^{(g)}_1)^2}{\sum_{g=1}^Gn_g},\\
		\widehat{\sigma}_{22} & = \frac{\sum_{g=1}^G\left[\sum_{j=1}^{m_g}(x^{(g)}_{2j}-\widehat{\mu}^{(g)}_2)^2+ \sum_{j=n+1}^{l_g}(x^{(g)}_{2j}-\widehat{\mu}^{(g)}_2)^2\right]}{\sum_{g=1}^G(m_g+l_g-n_g)},	
	\end{align*}
 
	and $\widehat{\sigma} _{12}$, where $\widehat{\sigma} _{12}$ is the maximizer of:	
 
	\begin{equation*}\label{etacov}
		\eta = C-\frac{A\log \left(\sigma_{22}-\frac{\sigma_{12}^2}{\sigma_{11}}\right)}{2}
		- \frac{\left( s_{22} - 2\frac{\sigma_{12}}{\sigma_{11}} s_{12}+\frac{\sigma_{12}^2}{\sigma _{11}^2}s_{11}\right)}{2\left(\sigma_{22}-\frac{\sigma_{12}^2}{\sigma_{11}}\right)}
	\end{equation*}
\end{theorem}
The estimate for $\mu^{(g)}_1, \sigma_{11},\mu^{(g)}_2, \sigma_{22}$ are explicitly formulated, but the estimation of $\sigma_{12}$ is implicit, and the following theorem provides a way to estimate $\sigma_{12}$.
\begin{theorem}\label{theorem:root}
    The estimation of $\sigma_{12}$ can be reduced to finding the roots of the following polynomial
    \begin{equation}\label{eq:sigma}
        P(\sigma_{12})=s_{12}\sigma_{11}\sigma_{22}+(\sigma_{11}\sigma_{22}A - s_{22}\sigma_{11}-s_{11}\sigma_{22})\sigma_{12} + s_{12}\sigma_{12}^2-A\sigma_{12}^3
    \end{equation}
    provided that 
    \begin{equation*}
         s_{12} \neq \pm \frac{s_{22}\sigma_{11}+s_{11}\sigma_{22}}{2\sqrt{\sigma_{11}\sigma_{22}}}\label{eq:cond roots}
    \end{equation*}
\end{theorem}
Based on these theoretical grounds, Algorithm~\ref{alg:DPERm} provides a way to directly estimate parameters for multiple-class data under the assumption of equal covariance matrices.
\begin{algorithm}
\raggedright
\caption{\textbf{DPER for multiple-class data}}\label{alg:DPERm}
    \hspace*
    {\algorithmicindent} 
    \textbf{Input:} A dataset $\mathbf{X}$ has $p$ continuous features  $\mathbf{f}_1,\mathbf{f}_2,...,\mathbf{f}_p$ and $\mathbf{y}$ as labels with $G$ classes.\\
    \hspace*{\algorithmicindent} 
    \textbf{Output:} the covariance matrix estimation $\widehat{\boldsymbol{\Sigma}} = \left( \widehat{\sigma}_{ij} \right)_{i,j=1}^p$.
\begin{algorithmic}[1]
    \State Estimate $\widehat{\boldsymbol{\mu}}^{(g)}$: $\widehat{\mu}_i^{(g)} \leftarrow$  mean of all observations of class $g$ in $\mathbf{f}_i$.
    \State $\left( \widehat{\sigma}_{ii} \right)_{i=1}^p \leftarrow$ uncorrected sample variance of all observations in $\mathbf{f}_i$
    \For {$1\le i \le p$}
        \For {$1\le j \le i$}
                \State $\widehat{\sigma}_{ij}=\widehat{\sigma}_{ji} \leftarrow$ root of the polynomial \eqref{eq:sigma}, closest to case deletion estimation
        \EndFor 
    \EndFor
\end{algorithmic}
\end{algorithm}

In case $G=1$, in other words, the data is single-class data, then one can have the following theorems as the corollaries of Theorems~\ref{bivariatemle_multi} and~\ref{theorem:root}, respectively. Moreover, the corresponding algorithm will also be provided.
\begin{theorem}\label{bivariatemle_single}
	Assume that we have a set of i.i.d observations from a bivariate normal distribution with mean  $
		\boldsymbol {\mu } = \begin{pmatrix}
			\mu_1\\ \mu_2
		\end{pmatrix} $
	and covariance matrix $
		\boldsymbol {\Sigma } =  \begin{pmatrix}
			\sigma_{11}&\sigma_{12}\\
			\sigma _{21} & \sigma _{22}
		\end{pmatrix},$
	and arrange the data into the following pattern
	\begin{equation*}\label{bivariate-form}
		\boldsymbol{x} = \begin{pmatrix}
			{x}_{11} & ... & {x}_{1m}&{x}_{1m+1}&...&{x}_{1n}&*&...&*\\
			{x}_{21}&...&{x}_{2m}& *& ...& *& {x}_{2n+1}&...&{x}_{2l}
		\end{pmatrix}.
	\end{equation*}
	So, each column represents an observation, and $x_{ij} \in \mathbb{R}$ is an entry, i.e., each observation has two features.  
	
	Let $L$ be the likelihood of the data and	
	\begin{align*}
		s_{11} &= \sum_{j=1}^{m}(x_{1j}-\hat{\mu}_1)^2	,\\
		s_{12} &= \sum_{j=1}^{m}(x_{2j}-\hat{\mu}_2)(x_{1j}-\hat{\mu}_1),\\
		s_{22} &= \sum_{j=1}^{m}(x_{2j}-\hat{\mu}_2)^2.
	\end{align*}
	
	Then, the resulting estimators obtained by maximizing $L$ w.r.t $\mu_1, \sigma_{11},\mu_2, \sigma_{22}$, and $\sigma_{12}$ are:
 
	\begin{align*}
		\hat{\mu}_1 & = \frac{1}{n}\sum_ {j=1}^{n}x_{1j},\;\;
		\hat{\mu}_2 = \frac{\sum_ {j=1}^{m}x_{2j}+ \sum_ {j=n+1}^{l}x_{2j}}{m+l-n},	 \\
		\hat{\sigma}_{11} & = \frac{\sum_ {j=1}^n(x_{1j}-\hat{\mu}_1)^2}{n},\\
		\hat{\sigma}_{22} & = \frac{\sum_ {j=1}^{m}(x_{2j}-\hat{\mu}_2)^2+ \sum_ {j=n+1}^{l}(x_{2j}-\hat{\mu}_2)^2}{m+l-n},	
	\end{align*}
	
	and $\hat{\sigma} _{12}$, where $\hat{\sigma} _{12}$ is the maximizer of:	
	\begin{equation*}\label{etacov-sing}
		\eta = C-\frac{m\log \left(\sigma_{22}-\frac{\sigma_{12}^2}{\sigma_{11}}\right)}{2}
		- \frac{\left( s_{22} - 2\frac{\sigma_{12}}{\sigma_{11}} s_{12}+\frac{\sigma_{12}^2}{\sigma _{11}^2}s_{11}\right)}{2\left(\sigma_{22}-\frac{\sigma_{12}^2}{\sigma_{11}}\right)}.
	\end{equation*}
\end{theorem}
\begin{theorem}
 The estimation of $\sigma_{12}$ can be reduced to finding the roots of the following polynomial
    \begin{equation}\label{eq:sigma-sing}
       P(\sigma_{12})=s_{12}\sigma_{11}\sigma_{22}+(\sigma_{11}\sigma_{22}A - s_{22}\sigma_{11}-s_{11}\sigma_{22})\sigma_{12} + s_{12}\sigma_{12}^2-m\sigma_{12}^3
    \end{equation}
    provided that 
    \begin{equation*}
         s_{12} \neq \pm \frac{s_{22}\sigma_{11}+s_{11}\sigma_{22}}{2\sqrt{\sigma_{11}\sigma_{22}}}.
    \end{equation*}   
\end{theorem}

\begin{algorithm}
\raggedright
\caption{\textbf{DPER for single-class data}}\label{alg:DPER}
    \hspace*{\algorithmicindent} \textbf{Input:} A dataset $\mathbf{X}$ has $p$ continuous features $\mathbf{f}_1,\mathbf{f}_2,...,\mathbf{f}_p$.\\
    \hspace*{\algorithmicindent} \textbf{Output:} the covariance matrix estimation $\widehat{\boldsymbol{\Sigma}} = \left( \widehat{\sigma}_{ij} \right)_{i,j=1}^p$.
\begin{algorithmic}[1]
    \State Estimate $\boldsymbol{\mu}$: $\mu_i \leftarrow$  mean of all observations in $\mathbf{f}_i$.
    \State $\left( \widehat{\sigma}_{ii} \right)_{i=1}^p \leftarrow$ uncorrected sample variance of all observations in $\mathbf{f}_i$
    \For {$1\le i \le p$}
        \For {$1\le j \le i$}
                \State $\widehat{\sigma}_{ij}=\widehat{\sigma}_{ji} \leftarrow$ solve the equation \eqref{eq:sigma-sing}
                \If {equation \eqref{eq:sigma-sing} has more than one root} chose one closest to case deletion estimation
                \EndIf
        \EndFor 
    \EndFor
\end{algorithmic}
\end{algorithm}
 Note that the proof of all theorems above can be found in~\cite{NGUYEN2022108082}.

\section{DPERC algorithm for single-class and multiple-class mixed data}\label{sec:DPERC single}

In this section, we aim to present a new algorithm, DPERC, for direct parameter estimating single-class and multiple-class mixed data.

\subsection{Fundamental idea and theoretical grounds}\label{subsec: Fundamental}
We assume $\mathbf{X=(X_{cont}, X_{cate})}$ is a single class mixed-type data, where $\mathbf{X_{cont}}$ has $p$ features and $\mathbf{X_{cate}}$ has $q$ features $\mathbf{c}_1, \mathbf{c}_2, \dots, \mathbf{c}_q$. The DPER algorithm only uses the information from continuous features in $\mathbf{X_{cont}}$ and ignores the categorical features in $\mathbf{X_{cate}}$. This means information from the categorical features, which could be useful, is not utilized. Therefore, we propose to use the categorical features as class features under the assumption of equal covariance matrices in estimating the covariance matrices. In particular, to estimate the covariance matrix $\boldsymbol{\Sigma}$ of $\mathbf{X_{cont}}$, we treat each feature $\mathbf{c}_i$ of $\mathbf{X_{cate}}$ as a new artificial class feature where $i=1,2,\dots,q$. Moreover, suppose that $\mathbf{c_i}$ has $G_i$ categories, then we assume that $\boldsymbol{\Sigma}^{(1)} = \boldsymbol{\Sigma}^{(2)} = \dots = \boldsymbol{\Sigma}^{(G_i)}$ where $\boldsymbol{\Sigma}^{(g)}$ for $g=1,2,\dots,G_i$, is the covariance matrix of the $g^{th}$ class which has $\mathbf{c}_i=g$. Then, we use the DPER multiple-class algorithm for data $(\mathbf{X_{cont}},\mathbf{c}_i)$. Obviously, this can result in $q$ estimated covariance matrix. So we want to find a condition to choose the good one from those. Here, it means we choose a better estimate of the covariance matrix than the result of DPER, i.e., 
\begin{equation*}
    \|\widehat{\boldsymbol{\Sigma}}_{DPERC} - \boldsymbol{\Sigma}\|_F \le \|\widehat{\boldsymbol{\Sigma}}_{DPER} - \boldsymbol{\Sigma}\|_F,
\end{equation*}
where $\|.\|_F$ is the Frobenius norm. 

This signifies we need a condition to choose a categorical feature $\mathbf{c}_i$ for good performances. Going into detail with data of 2 features, assume that we have a data $\boldsymbol{x}$ and a categorical feature $\mathbf{c}$, which has $G$ categories $1,2,\dots, G$. i.e.
    \begin{align*}
        \boldsymbol{x} &= \begin{pmatrix}
            x_{11} & x_{12} & \dots & x_{1N} \\
            x_{21} & x_{22} & \dots & x_{2N} \\
        \end{pmatrix},
    \end{align*}
where $\mathbf{u}_{j}^{(g)} = (x_{1j}^{(g)},x_{2j}^{(g)})$ belong to $g^{th}$ category and suppose that the number of observations of the $g^{th}$ category is $n_{g}$, for $1\le g \le G$. And also for $1\le g \le G$, we first denote the \textbf{weighted average pairwise distance} of the $g^{th}$ category of data $\boldsymbol{x}$ follow $\mathbf{c}$
\begin{align*}
    d^{(g)}_{\mathbf{c}} = \sum_{1\le i<j\le n_g}\frac{1}{n_g}(\mathbf{u}_i^{(g)}-\mathbf{u}_{j}^{(g)})^T\boldsymbol{\Sigma}^{-1}(\mathbf{u}_i^{(g)}-\mathbf{u}_{j}^{(g)}),
\end{align*}
where $\boldsymbol{\Sigma}$ is the covariance matrix of data. In particular, $d^{(g)}_{\mathbf{c}}$ presents the average of the distances between any two elements of data $\boldsymbol{x}$.

\begin{theorem}\label{theorem:dperc}
Suppose that we have a single-class data $\boldsymbol{x}$ from
a bivariate normal distribution with mean $\boldsymbol{\mu}$ and covariance matrix $\boldsymbol{\Sigma}$, and a categorical feature $\mathbf{c}$ which has $G$ categories $1,2,\dots, G$. i.e.
    \begin{align*}
        \boldsymbol{x} &= \begin{pmatrix}
            x_{11} & x_{12} & \dots & x_{1N} \\
            x_{21} & x_{22} & \dots & x_{2N} \\
        \end{pmatrix}.
    \end{align*}
Moreover, the observations of the $g^{th}$ category be $\mathbf{u}_{1}^{(g)},\mathbf{u}_{2}^{(g)},\dots,\mathbf{u}_{n_g}^{(g)}$ with mean $\boldsymbol{\mu}^{(g)}$, where $1\le g\le G$. Also for $1\le g\le G$, we denote
\begin{align*}
\delta^{(g)} &= \sum^{n_g}_{i=1}(\mathbf{u}_{i}^{(g)}-\boldsymbol{\mu})^T\boldsymbol{\Sigma}^{-1}(\mathbf{u}_{i}^{(g)}-\boldsymbol{\mu}),\\
\Delta^{(g)} &= (\boldsymbol{\mu}^{(g)}-\boldsymbol{\mu})^T\boldsymbol{\Sigma}^{-1}(\boldsymbol{\mu}^{(g)}-\boldsymbol{\mu}).
\end{align*}
Then 
\begin{equation}
    \delta^{(g)}=d^{(g)}_{\mathbf{c}}+ n_g\Delta^{(g)},\quad g = 1,2,\dots,G.
\end{equation}
\end{theorem}
The proof of this theorem can be found in the Appendix~\ref{append-proof}. Now, also in~\cite{NGUYEN2022108082}, the log-likelihood of data $\boldsymbol{x}$ is 
\begin{equation*}
   L = C - \frac{N}{2}\log|\boldsymbol{\Sigma}| - \frac{1}{2}\sum_{i=1}^{N}(\mathbf{u}_{i}-\boldsymbol{\mu})^T\boldsymbol{\Sigma}^{-1}(\mathbf{u}_{i}-\boldsymbol{\mu}).
\end{equation*}

The last term can be rewritten as
\begin{align*}
  \sum_{i=1}^{N}(\mathbf{u}_{i}-\boldsymbol{\mu})^T\boldsymbol{\Sigma}^{-1}(\mathbf{u}_{i}-\boldsymbol{\mu}) \nonumber =\sum^G_{g=1}\sum^{n_g}_{j=1}(\mathbf{u}_{j}^{(g)}-\boldsymbol{\mu})^T\boldsymbol{\Sigma}^{-1}(\mathbf{u}_{j}^{(g)}-\boldsymbol{\mu}) =\sum_{g=1}^{G}\delta^{(g)}.
\end{align*}
So, we can rewrite $L$ to
\begin{equation*}
    L = C - \frac{N}{2}\log|\boldsymbol{\Sigma}| - \frac{1}{2}\sum_{g=1}^{G}\delta^{(g)}.
\end{equation*}
By the Theorem~\ref{theorem:dperc},
\begin{equation*}
    L = C - \frac{N}{2}\log|\boldsymbol{\Sigma}| - \frac{1}{2}\sum_{g=1}^{G}(d^{(g)}_{\mathbf{c}}+n_g\Delta^{(g)}).
\end{equation*}
We want to maximize $L$ by minimize $\sum_{g=1}^{G}(d^{(g)}_{\mathbf{c}}+n_g\Delta^{(g)})$. 
Here, we could control the value of $n_g\Delta^{(g)}$ by $\mathbf{c}$. It follows that we can choose the categorical feature $\mathbf{c}$ so that $\sum_{g=1}^{G}n_g\Delta^{(g)}$ is minimal.\\
Intuitively, if the sum of average pairwise distances between all samples in a group is small, for each class, the observations are clustered together in one cluster. Therefore, in such a case, one can expect the sum of the distances from all samples to the central mean to be small if the sum of distances of all class means to the central mean is small. In other words, we choose the categorical feature $\mathbf{c}$ such that the elements in the group of category $g^{th}$ do not have much difference. 
That is the standard for choosing a good categorical feature. As mentioned previously, after having a good categorical feature $\boldsymbol{c}$ for data $\boldsymbol{x}$, we treat this feature as a class feature and then use the DPER algorithm for multiple-class to estimate the covariance matrix.\\
\subsection{DPERC for single-class mixed data}\label{subsec: DPERc algorithm}
\begin{algorithm}
\raggedright
\caption{\textbf{DPERC for single-class mixed data}}\label{alg:DPERC}
\textbf{Input:} A data $\mathbf{X=(X_{cont},X_{cate})}$ where $\mathbf{X_{cont}}$ has $p$ continuous features $\mathbf{f}_1,\mathbf{f}_2,...,\mathbf{f}_p$ and $\mathbf{X_{cate}}$ has $q$ categorical features $\mathbf{c}_1,\mathbf{c}_2,\dots,\mathbf{c}_q$.\\
\textbf{Output:} the covariance matrix estimate $\widehat{\boldsymbol{\Sigma}} = \left( \widehat{\sigma}_{ij} \right)_{i,j=1}^p$.\\
\textbf{Procedure:} 
\begin{algorithmic}[1]
    \State Estimate $\widehat{\boldsymbol{\mu}}$: $\widehat{\mu_i} \leftarrow$  mean of all observations in $\mathbf{f}_i$
    \State $\widehat{\boldsymbol{\Sigma}}_{D} \leftarrow$ using DPER for single-class $\mathbf{X_{cont}}$ data
    \State $\left( \widehat{\sigma}_{ii} \right)_{i=1}^p \leftarrow$ uncorrected sample variance of all observations in $\mathbf{f}_i$
    \For {$1\le i \le p$}
        \For {$1\le j \le i$}
            \For{$1 \le k \le q$}
                \State $\mathcal{D}_{\boldsymbol{c_k}} \leftarrow \sum^{\# \mathbf{c_k}}_{g=1}n_g(\widehat{\boldsymbol{\mu}}_{ij}^{(g)}-\widehat{\boldsymbol{\mu}}_{ij})^T \widehat{\boldsymbol{\Sigma}}_{D,ij}^{-1}(\widehat{\boldsymbol{\mu}}_{ij}^{(g)}-\widehat{\boldsymbol{\mu}}_{ij})$
            \EndFor
            \State $\mathcal{D}_{\boldsymbol{c}} \leftarrow \{\mathcal{D}_{\boldsymbol{c_1}}, \mathcal{D}_{\boldsymbol{c_2}}, \dots, \mathcal{D}_{\boldsymbol{c_q}} \}$
            \State $\mathbf{c} \leftarrow \arg\min_{\boldsymbol{c}} 
            \mathcal{D}_{\boldsymbol{c}}$
            \State $\widehat{\sigma}_{ij}=\widehat{\sigma}_{ji} \leftarrow$ using DPER for multiple-class on $(\mathbf{f}_i,\mathbf{f}_j)$ and $\mathbf{c}$ as a class
        \EndFor 
    \EndFor
\end{algorithmic}
\end{algorithm}

In this section, we will detail the DPERC algorithm for estimating the covariance matrix for single-class mixed data. To start, we have some notations. Note that $\widehat{\boldsymbol{\mu}}$ is the estimated mean of data. Recall that for each categorical feature $\boldsymbol{c}$ which has $\#\boldsymbol{c}$ different labels, we treat it as a class feature, so this separates the data $\mathbf{X_{cont}}$ into $\#\boldsymbol{c}$ class data. Therefore, we denote $ \widehat{\boldsymbol{\mu}}^{(g)}$ is the estimated mean of $g^{th}$ class data where $1\le g\le \#\boldsymbol{c}$. Next, denote $\widehat{\boldsymbol{\mu}}_{ij}, \widehat{\boldsymbol{\mu}}^{(g)}_{ij}$ for a vector consists $i^{th}$ and $j^{th}$ elements of $\widehat{\boldsymbol{\mu}}$ and $\widehat{\boldsymbol{\mu}}^{(g)}$ respectively, i.e., if 
$\widehat{\boldsymbol{\mu}} = (m_1,m_2,m_3,m_4)$, where $m_i \in \mathbb{R}$ then $\widehat{\boldsymbol{\mu}}_{13} = (m_1,m_3)$.\\
In addition, $\widehat{\boldsymbol{\Sigma}}^{-1}_{D,ij} = \begin{pmatrix}
        \lambda_{ii} & \lambda_{ij} \\
        \lambda_{ji} & \lambda_{jj} 
    \end{pmatrix} $ where $\lambda_{ii}, \lambda_{jj}, \lambda_{ij}, $ and $\lambda_{ji}$ are the corresponding elements of $\widehat{\boldsymbol{\Sigma}}_{D}^{-1}$, and $\widehat{\boldsymbol{\Sigma}}_{D}$ is the estimated covariance matrix by using DPER.

The DPERC algorithm for single-class mixed data is presented in Algorithm \ref{alg:DPERC}. The algorithm input requires the mixed data and returns its covariance matrix estimation, and the procedure is as follows, 
\begin{enumerate}[label=(\alph*)]
    \item Step 1: we estimate the mean vector by calculating the mean of each feature after removing any missing values (case deletion).
    \item Step 2: we can use another technique to estimate the covariance matrix $\widehat{\Sigma}_D$. This one is denoted as a first covariance matrix estimate, which is utilized to choose the good categorical feature.
    \item Step 3: we estimate the diagonal elements of the estimated covariance matrix by uncorrected sample variance (i.e., the denominator is divided by the number of observed elements).
    \item In the next steps, we focus on estimating the elements outside the diagonal, i.e., $\sigma_{ij} (i\neq j)$. From step 7 to step 9, we compute the \textbf{weighted average pairwise distance} follow $\boldsymbol{c}$ and then put all these distances in a set $\mathcal{D}_{\boldsymbol{c}}$.   
    \item In step 10, we choose the categorical feature $\boldsymbol{c}$ such as this minimize the weighted average pairwise distance, i.e., finding the minimal value in the set $\mathcal{D}_{\boldsymbol{c}}$. 
    \item In the last step 11, we use DPER for multiple-class to estimate $\sigma_{ij}$.
\end{enumerate}

There are some remarks about the algorithm
\begin{enumerate}[label=(\alph*)]
    \item Similar to DPER, we also need a mild assumption of bivariate normality on each pair of continuous features. Moreover, we need to go through all the categorical features for each pair $(\mathbf{f}_i,\mathbf{f}_j)$ to choose the good one for estimating each $\sigma_{ij}$. However, one more point to note, we do not estimate the first covariance matrix for each pair of features. Instead, we use the corresponding principal sub-matrix of $\widehat{\Sigma}_D$ in step 2. 
    \item Note in step 7, when the missing rate is high, if the element $i^{th}$ of  $\widehat{\boldsymbol{\mu}}^{(g)}$ is missing, we set this element to zero.
    \item We can call this process a two-step estimation; the first estimation is considered as a factor in choosing the categorical feature for the second estimation.
\end{enumerate}

\subsection{DPERC algorithm for multiple-class data}\label{sec: DPERC multiple}
In case data comes from multiple classes without the assumption of equal covariance matrices, one can apply Algorithm~\ref{alg:DPERC} multiple times as the multiple classes problem boils down to many single-class problems.

For instance, suppose that we have a  mixed data $\mathbf{(X_{cont},X_{cate},y)}$ where $\mathbf{y}$ has $M$ classes. Let $\mathbf{X}^{(m)}$  be the data from the $m^{th}$ class and  $\mathbf{X}^{(m)}_{\mathbf{cont}}$ be its corresponding continuous part.
For each class $m$ where $1\le m\le M$, we want to estimate the covariance matrix of this class without the assumption of equal covariance matrices. 

Now, we denote $\mathbf{X_{c}}$ as a feature of $G$ categories from $\mathbf{X_{cate}}$. Then, we assume $\boldsymbol{\Sigma}_1^{(m)} = \boldsymbol{\Sigma}_2^{(m)} = \dots = \boldsymbol{\Sigma}_G^{(m)}$ where $\boldsymbol{\Sigma}^{(m)}_i$ is the covariance matrix of the $m^{th}$ class, i.e., the part that has$\mathbf{X}_{\mathbf{c}}^{(m)}=i$. Therefore, we can use Algorithm \ref{alg:DPERC} for $(\mathbf{X}^{(m)}_{\mathbf{cont}},\mathbf{X}^{(m)}_{\mathbf{cate}},\mathbf{y}^{(m)})$ to estimate the covariance matrix of class $m^{th}$. 
 
\section{Experiments}\label{sec:experiments}

This section will present our experiments related to the proposed algorithms and relevant discussions related to the experimental results.

\subsection{Experimental settings and datasets}

To illustrate the efficiency of our algorithm focused on the estimating covariance matrices task, we set up two comparisons. First, we compare the results of the proposed method to state-of-the-art methods, such as DPER~\cite{NGUYEN2022108082}, missForest~\cite{stekhoven2012missforest}, KNNI~\cite{beretta2016nearest}, Soft-Impute~\cite{mazumder2010spectral}, and MICE~\cite{buuren2010mice} 
with the similar metric as proposed in~\cite{NGUYEN2022108082}:
\begin{equation}\label{error e}
    e =\frac{\| \widehat{\boldsymbol{\Sigma}} - \boldsymbol{\Sigma}\|_F}{n_{\boldsymbol{\Sigma}}}
\end{equation}
where $\|.\|_F$ is Frobenius norm and $\widehat{\boldsymbol{\Sigma}}$ is the estimate of the ground truth $\boldsymbol{\Sigma}$ and $n_{\boldsymbol{\Sigma}}$ is the number of entries in $\boldsymbol{\Sigma}$.\\
Next, to illustrate the improvement of DPERC over DPER, note that both DPER and DPERC have the same estimate in diagonal elements of $\boldsymbol{\Sigma}$. Thus, we compare the estimate of $\sigma_{ij}$ where $i\neq j$. We use the following metric as in~\cite{nguyen4260235pmf}:

\begin{equation}\label{error r}
    r = \| \widehat{\boldsymbol{\Sigma}}^* - \boldsymbol{\Sigma}^*\|_F
\end{equation}
where $\widehat{\boldsymbol{\Sigma}}^*$ is the estimate of the ground truth $\boldsymbol{\Sigma}^*$, except the diagonal elements.\\
We evaluate the percentage improvement of DPERC compared to DPER by using the criterion:
\begin{equation}\label{percent}
 p = 100\%\left(1 - \frac{r_{DPERC}}{r_{DPER}}\right)
\end{equation}

To investigate the effects of missing data on the correlation heatmap, we employ three types of heatmap plots as proposed in \cite{pham2024correlation}:
\begin{enumerate}[label=(\alph*)]
    \item Correlation Heatmap illustrates the correlation matrix and employs a color gradient to illustrate correlation coefficients, ranging from -1 to 1. To convert a covariance matrix $\boldsymbol{\Sigma}$ to a correlation matrix $\boldsymbol{R}$, we use the following formula:
    \begin{equation}\label{cov_to_corr}
        \boldsymbol{R} = \boldsymbol{D}^{-1} \boldsymbol{\Sigma} \boldsymbol{D}^{-1}
    \end{equation}
    where $\boldsymbol{D}$ is the diagonal matrix where the $i$-th diagonal element is the standard deviation of the $i$-th feature, i.e., $\sqrt{\Sigma_{ii}}$.
    \item Local MSE Difference Heatmaps for Correlation provide a detailed comparison between the ground truth correlation matrix and the correlation matrices produced by different estimation techniques, offering visual insights into their discrepancies. By highlighting the squared differences for each element, these heatmaps pinpoint areas with significant variations. The computation of the Local MSE Difference matrix follows the formula:
    \begin{equation}
        \Delta \boldsymbol{R}_{\text{squared}} = \Delta \boldsymbol{R} \circ \Delta \boldsymbol{R}
    \end{equation}
    where $\circ$ denotes the element-wise product, $\Delta \boldsymbol{R}$ is the matrix containing the local differences between corresponding elements of the ground truth correlation matrix $\boldsymbol{R}_{\text{true}}$ and the correlation matrix produced by an estimation technique $\boldsymbol{R}_{\text{est}}$, $\Delta \boldsymbol{R}$ = $\boldsymbol{R}_{\text{true}} - \boldsymbol{R}_{\text{est}}$.
    \item Local Difference (Matrix Subtraction) Heatmaps for Correlation highlight differences between correlation matrices generated from each estimation technique and the ground truth correlation matrix. By emphasizing whether the differences are positive or negative, these heatmaps provide a clear indication of the direction of variation. This facilitates the identification of areas where estimation techniques either overestimate or underestimate correlations in comparison to the ground truth.
\end{enumerate}

The experiments are run directly on Google Colaboratory \footnote{\url{https://colab.google/}}, each experiment is repeated ten times, and we report the mean of error as mentioned in Equations \eqref{error e} and \eqref{error r} and percentage improvement (Equation \eqref{percent}). The datasets for all experiments are from the Machine Learning Database Repository at the University
of California \footnote{\url{https://archive.ics.uci.edu/ml}}: Bank Marketing, Statlog, Heart Disease, and Student Performance. More detail of these data is shown in Table~\ref{tab:data}. 
\begin{table}[!htp]
\centering
\begin{tabular}{|c|c|c|c|c|c|}
\hline
Dataset & \# Features & \begin{tabular}[c]{@{}c@{}}\# Continuous \\ features\end{tabular} & \begin{tabular}[c]{@{}c@{}}\# Categorical\\ features\end{tabular} & \# Classes & \# Samples\\
\hline
Bank & 21 & 10 & 10 & 2 & 4168 \\
Heart & 15 & 6 & 8 & 2 & 32561 \\
Statlog & 21 & 7 & 13 & 2 & 1000 \\
Student & 33 & 16 & 17 & 1 & 395 \\
\hline
\end{tabular}
\caption{Dataset description}\label{tab:data}
\end{table}

Moreover, we only consider missing values in continuous features; the categorical features are fully observed. We simulate missing data for every dataset with missing rates ranging from 20\% to 80\%. Here, the missing rate is defined as the proportion of missing entries to the total number of entries. The link to the source codes will be provided upon acceptance of the paper.

\subsection{Results and analysis}

\begin{table}
\centering
\begin{tabular}{|c|c|c|c|c|c|c|c|}
\hline
\textbf{Datasets}
& \parbox{1.3cm}{\textbf{Missing rate (\%)}} & \textbf{DPERC} & \textbf{DPER} & \textbf{missForest} & \textbf{KNN} & \textbf{Soft-Impute} & \textbf{MICE} \\
\hline
\multirow{5}{4em}{\textbf{Student}} 
& 20\% & \textbf{0.590} & 0.592 & 1.384 & 0.860 & 1.131 & 0.729 \\
& 35\% & \textbf{0.813} & 0.818 & 2.076 & 1.223 & 1.780 & 1.021 \\
& 50\% & \textbf{1.086} & 1.088 & 2.473 & 1.738 & 2.322 & 1.345 \\
& 65\% & \textbf{1.345} & 1.351 & 2.803 & 2.269 & 2.834 & 1.592 \\
& 80\% & \textbf{1.616} & 1.623 & 3.115 & 2.689 & 3.227 & 1.945 \\
\hline
\multirow{5}{4em}{\textbf{Bank}} 
& 20\% & \textbf{0.152} & 0.157 & 0.209 & 0.223 & 0.332 & 0.252 \\
& 35\% & \textbf{0.199} & 0.202 & 0.447 & 0.535 & 0.571 & 0.407 \\
& 50\% & \textbf{0.220} & 0.225 & 1.230 & 0.801 & 0.836 & 0.659 \\
& 65\% & \textbf{0.336} & 0.339 & 1.433 & 1.041 & 1.070 & 0.771 \\
& 80\% & \textbf{0.482} & 0.488 & 1.576 & 1.030 & 1.235 & 0.974 \\
\hline
\multirow{5}{4em}{\textbf{Statlog}}
& 20\% & \textbf{0.097} & 0.098 & 0.163 & 0.174 & 0.220 & 0.175 \\
& 35\% & \textbf{0.147} & 0.149 & 0.276 & 0.275 & 0.345 & 0.254 \\
& 50\% & \textbf{0.175} & 0.179 & 0.443 & 0.335 & 0.437 & 0.276 \\
& 65\% & \textbf{0.228} & 0.231 & 0.544 & 0.403 & 0.539 & 0.350 \\
& 80\% & \textbf{0.300} & 0.306 & 0.601 & 0.453 & 0.606 & 0.408 \\
\hline
\multirow{5}{4em}{\textbf{Heart}}
& 20\% & \textbf{0.076} & \textbf{0.076} & 0.156 & 0.153 & 0.191 & 0.190 \\
& 35\% & \textbf{0.102} & \textbf{0.102} & 0.274 & 0.264 & 0.325 & 0.322 \\
& 50\% & \textbf{0.060} & \textbf{0.060} & 0.418 & 0.400 & 0.480 & 0.477 \\
& 65\% & \textbf{0.109} & \textbf{0.109} & 0.514 & 0.455 & 0.574 & 0.556 \\
& 80\% & \textbf{0.088} & \textbf{0.088} & 0.609 & 0.572 & 0.683 & 0.632 \\
\hline
\end{tabular}
\caption{The estimation error based on the metric~\ref{error e}. The bold shows the best performance}\label{table2}
\end{table}

Table~\ref{table2} shows that DPERC performs best in all scenarios. Except for DPER, the estimation errors of DPERC have significantly improved compared to all other methods. Specifically, at the missing rate $80\%$, the result on the Statlog dataset, the error rate of DPERC is $0.3$, compared to the next best performer (MICE) at $0.408$.\\
We can also notice that the error rate increases in each case due to the missing rate. Yet, in most cases, the estimation error increment for DPERC is smaller than the other methods (except DPER). For instance, the error increment of DPERC on the Student dataset is $0.223-0.243-0.259-0.271$ while for missForest is $0.692-0.397-0.33-0.312$. Moreover, the improvement of DPERC compared to DPER is significant and necessary. For instance, in the Statlog dataset, when the missing rate is $35\%$, the improvement of DPERC compared to DPER is $0.002$ while the advantage of KNN is smaller than missForest, just $0.001$.

\begin{table}[!htp]
\centering
\begin{tabular}{|c|c|c|c|c|}
\hline
\textbf{Datasets}&
\parbox{1cm}{\textbf{Missing rate(\%)}}  &\textbf{DPERC} & \textbf{DPER} & \textbf{Improvement (\%)} \\

\hline
\multirow{5}{4em}{\textbf{Bank}} 
& 20\% & 0.312 & 0.332 & 6.1\% \\ 
& 35\% & 0.438 & 0.500 & 2.5\% \\ 
& 50\% & 0.601 & 0.613 & 2.0\% \\ 
& 65\% & 0.915 & 0.924 & 1.0\% \\
& 80\% & 1.272 & 1.289 & 1.3\% \\ 
\hline
\multirow{5}{4em}{\textbf{Heart}}
& 20\% & 0.033 & 0.033 & 0.7\% \\
& 35\% & 0.056 & 0.057 & 2.4\% \\
& 50\% & 0.078 & 0.080 & 2.6\% \\
& 65\% & 0.091 & 0.092 & 1.2\% \\
& 80\% & 0.117 & 0.117 & 0.4\% \\
\hline
\multirow{5}{4em}{\textbf{Statlog}}
& 20\% & 0.235 & 0.239 & 1.7\% \\
& 35\% & 0.380 & 0.388 & 2.2\% \\
& 50\% & 0.472 & 0.481 & 2.0\% \\
& 65\% & 0.621 & 0.630 & 1.4\% \\
& 80\% & 0.788 & 0.807 & 2.4\% \\
\hline
\multirow{5}{4em}{\textbf{Student}}
& 20\% & 0.397 & 0.398 & 0.3\% \\
& 35\% & 0.531 & 0.535 & 0.7\% \\
& 50\% & 0.727 & 0.729 & 0.2\% \\
& 65\% & 0.907 & 0.911 & 0.5\% \\
& 80\% & 1.086 & 1.091 & 0.5\% \\
\hline
\end{tabular}

\caption{The estimation error of DPER and DPERC based on the metric \eqref{error r} and the percentage of improvement \eqref{percent}.}
\label{table-percent}
\end{table}

Table~\ref{table-percent} shows that in most cases, the result of DPERC improves about $0.5\%$ to $2.5\%$ compared to DPER. Specifically, in the results on multiple classes of data, Bank and Statlog, there is at least a $1.0\%$ and $1.4\%$ improvement, respectively. In contrast, that worth in Heart's result is $0.4\%$. One explanation could be that the number of categorical features in Bank and Statlog is higher than that in Heart data, so one can choose the better one. Therefore, there is no surprise when DPERC only improved less than $1\%$ compared to DPER on the single class dataset (Student). Another noticeable feature is that the percentage improvement seems to be independent of the missing rate. For example, the percentage improvement at the missing rate $20\%$ is the highest in the result on the Bank dataset, while that worth in the result on the Heart dataset occurs at the missing rate $50\%$.

\subsection{Correlation visualization}
In this subsection, we analyze the correlation heatmaps derived from the covariance matrices estimated using the methods described in Section \ref{sec:experiments} on the Statlog dataset. We then compare correlation matrices produced by each estimation technique with the ground truth correlation matrix by evaluating the Local MSE Difference for Correlation and Local Difference (Matrix Subtraction) for Correlation. Each row in the subfigure represents a different missing ratio, ranging from 20\% to 80\%.

\begin{figure}[!htp]
    \centering
    \includegraphics[width=\columnwidth]{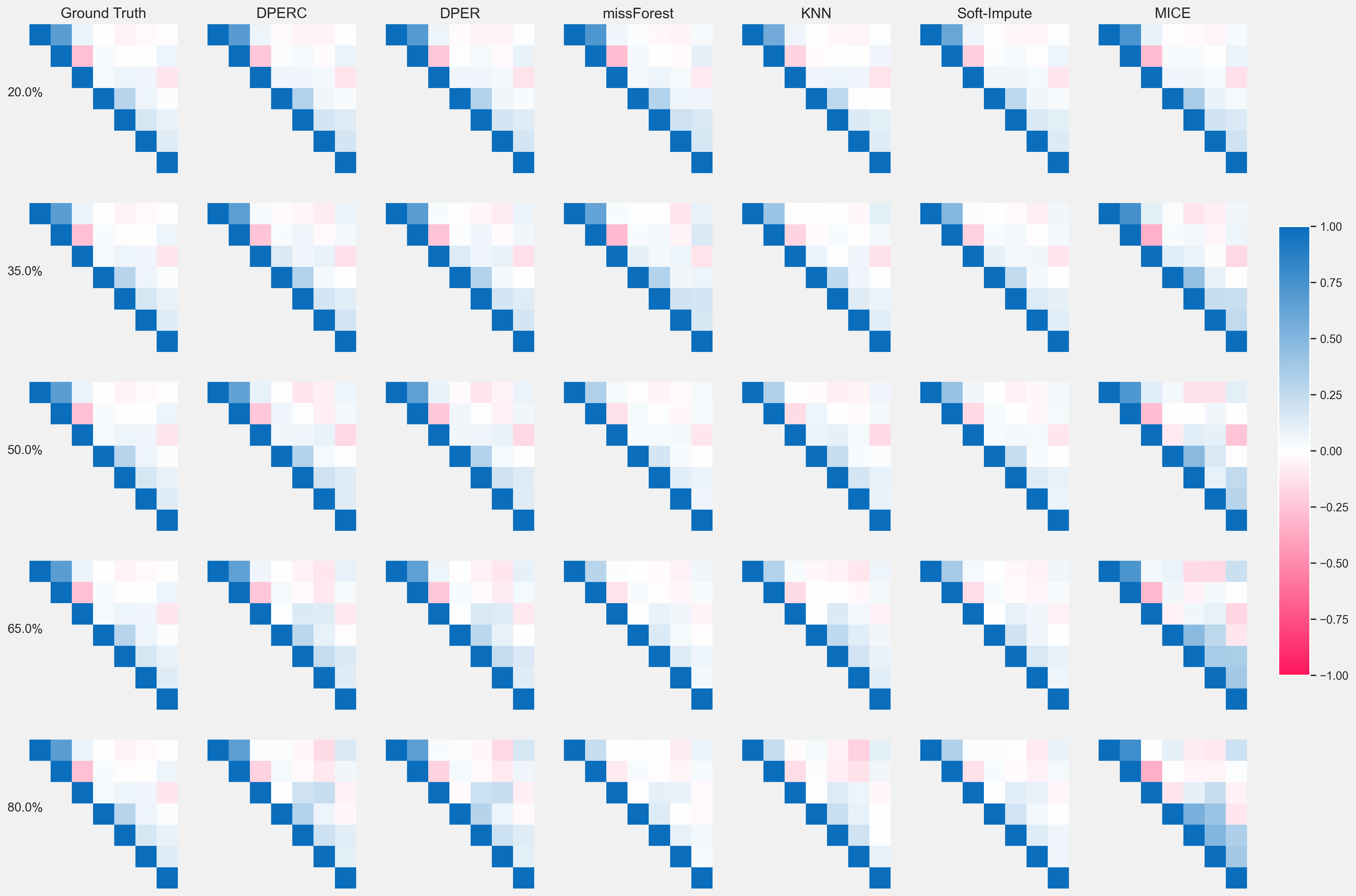}
    \caption{Correlation Heatmaps for the Statlog dataset across missing rates from 0.2 to 0.8.}
    \label{fig:corr_heatmap}
\end{figure}
\begin{figure}[!htp]
    \centering
    \includegraphics[width=\columnwidth]{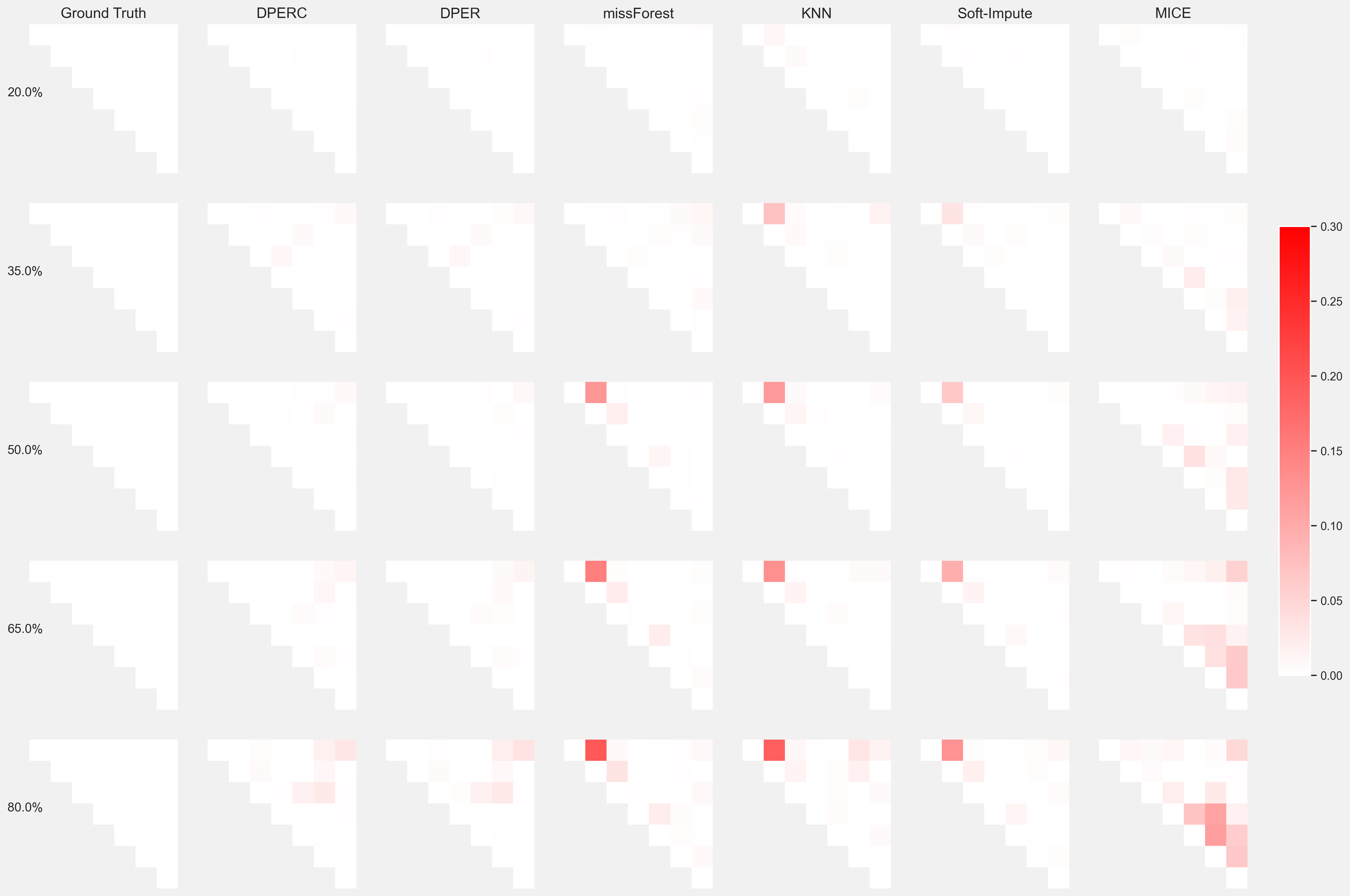}
    \caption{Local MSE Difference Heatmaps for Correlation for the Statlog dataset across missing rates from 0.2 to 0.8.}
    \label{fig:mse_corr}
\end{figure}
\begin{figure}[!htp]
    \centering
    \includegraphics[width=\columnwidth]{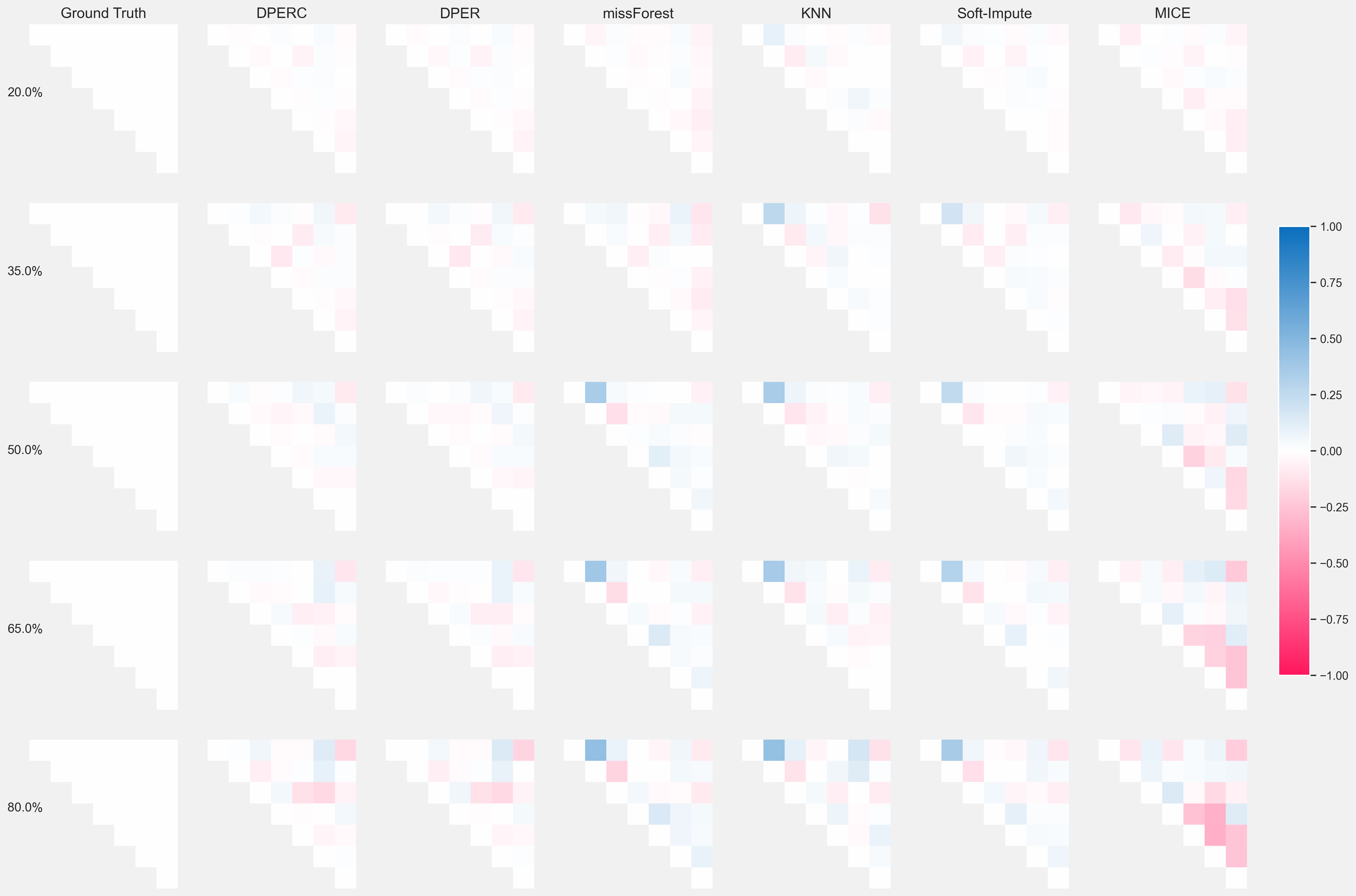}
    \caption{Local Difference (Matrix Subtraction) Heatmaps for Correlation for the Statlog dataset across missing rates from 0.2 to 0.8.}
    \label{fig:sub_corr}
\end{figure}

Figure \ref{fig:corr_heatmap} illustrates the correlation heatmaps, which visually represent the correlation between continuous features, with values ranging from -1 to 1. The negative correlations are depicted in warmer colors (magenta), and positive correlations in cooler colors (blue), and median (0) values are indicated by white. As the correlation level increases to 1, the corresponding color tones become closer to the coolest colors. As the correlation level decreases to -1, corresponding color tones become closer to the warmest colors. At a glance, the correlation heatmaps of all methods do not reveal clear differences among them. 

Figure \ref{fig:mse_corr} further enhances the clarity of these visualizations. The intensity of the red colors corresponds to the magnitude of the difference between the correlation matrix and the ground truth (depicted as white). Noticeably, as the missing rates increase, the differences become more pronounced. When the missing rate is at 65\%, the differences between DPERC, DPER, and ground truth are insignificant. In contrast, missForest, KNN, and Soft-Impute show substantial changes in some values compared to the ground truth, with the most substantial differences corresponding to cells with high correlation values, and these methods yield similar results. The MICE correlation matrix differs the most from the ground truth, but the differences are widespread rather than deep. 

Figure \ref{fig:sub_corr} reveals both positive and negative disparities. The blue color implies that the ground truth value is higher than the corresponding value of the methods compared, and the magenta color indicates the reverse. The darker the color, the greater the difference. This type of heatmap provides a clear indication of the direction of the difference. For example, at a 65\% missing rate, we observe that the missForest, KNN, and Soft-Impute methods underestimate (blue cells) high correlation values, which occurs similarly among these methods. In contrast, the MICE plot shows a prevalence of magenta colors with a noticeable intensity, indicating that MICE is overestimating the correlation, especially when the features are strongly correlated. Notably, the differences between DPERC, DPER, and the ground truth are in pastel shades, demonstrating their effectiveness in estimating correlation matrices due to negligible differences.

While Table \ref{table2} shows that DPERC is slightly better than DPER in terms of the average Frobenius norm (Equation \eqref{error e}) and the correlation comparison plots show comparable performance between DPERC and  DPER, Table \ref{table-percent}, which measure the evaluation criteria as mentioned in Equations \eqref{error r} and \eqref{percent} show that DPERC clearly achieves lower error compared DPER in terms of the cumulative Frobenius error norm $r=||\hat{\mathbf{\Sigma}}^*-\mathbf{\Sigma}||_F$ and criterion $p$ (Equation \eqref{percent}).

\section{Conclusion and future works}
\label{sec:conclusion}
In this article, we propose the algorithm DPERC to estimate the covariance matrix directly from mixed data. Because of relying on the DPER algorithm, our proposed algorithm not only keeps the strength of DPER but also efficiently exploits the categorical features by treating them as a class under the assumption of equal covariance matrices. In particular, the technique presents choosing a categorical feature that satisfies the Theorem~\ref{theorem:dperc} to improve the estimate for each $\sigma_{ij} (i\neq j)$. The experiments show that compared to DPER then, DPERC helps significantly enhance the covariance matrix estimation up to $6\%$, especially when data come from multiple classes. Furthermore, DPERC also achieves the lowest estimation error compared to other methods. Although the proposed method has some potential weaknesses, these will be interesting topics in the future.

For future works, first, note that our methodology provides a condition for choosing a good, not the best, categorical feature. This means we still need to find better conditions for choosing categorical features to improve DPERC.
Secondly, when dealing with data from multiple classes, we work with two assumptions: equal/non-equal covariance matrices. We have successfully tackled the challenge of non-equal covariance matrices, which paves the way for a potential expansion of our algorithm to multi-class data under the assumption of equal covariance matrices.
Next, our work only assumes that the data has randomly missing values in continuous features. When the labels of the class contain missing values, even if categorical features have missing values, the categorical features may be utilized to handle this issue. It can be our future topic.
Finally, with our strategy of choosing categorical feature $\mathbf{c}$, we assume that $n_g\Delta^{(g)}$ is considerably smaller than $d^{(g)}_{\mathbf{c}}$. However, we explore that $n_g\Delta^{(g)}$  can be much higher compared to $d^{(g)}_{\mathbf{c}}$ in some cases. It can lead to a future direction for better choosing categorical feature strategy.


\bibliographystyle{ACM-Reference-Format}
\bibliography{ref}
\appendix
\section{Proof of Theorem~\ref{theorem:dperc}} \label{append-proof}
Suppose that we have a single-class data $\boldsymbol{x}$ from
a bivariate normal distribution with mean $\boldsymbol{\mu}$ and covariance matrix $\boldsymbol{\Sigma}$, and a categorical feature $\mathbf{c}$ which has $G$ categories $1,2,\dots, G$. i.e.
    \begin{align*}
        \boldsymbol{x} &= \begin{pmatrix}
            x_{11} & x_{12} & \dots & x_{1N} \\
            x_{21} & x_{22} & \dots & x_{2N} \\
        \end{pmatrix}.
    \end{align*}
Moreover, the observations of the $g^{th}$ category be $\mathbf{u}_{1}^{(g)},\mathbf{u}_{2}^{(g)},\dots,\mathbf{u}_{n_g}^{(g)}$ with mean $\boldsymbol{\mu}^{(g)}$, where $1\le g\le G$. Also for $1\le g\le G$, we denote
\begin{align*}
\delta^{(g)} &= \sum^{n_g}_{i=1}(\mathbf{u}_{i}^{(g)}-\boldsymbol{\mu})^T\boldsymbol{\Sigma}^{-1}(\mathbf{u}_{i}^{(g)}-\boldsymbol{\mu}),\\
\Delta^{(g)} &= (\boldsymbol{\mu}^{(g)}-\boldsymbol{\mu})^T\boldsymbol{\Sigma}^{-1}(\boldsymbol{\mu}^{(g)}-\boldsymbol{\mu}).
\end{align*}
Then 
\begin{equation}
    \delta^{(g)}=d^{(g)}_{\mathbf{c}}+ n_g\Delta^{(g)},\quad g = 1,2,\dots,G.
\end{equation}

\textit{Poof.} Consider class $g^{th}$ of $\mathbf{c}$, expanding  $\delta^{(g)}$ we get
\begin{align*}
   \delta^{(g)} & = \sum^{n_g}_{i=1}(\mathbf{u}_{i}^{(g)}-\boldsymbol{\mu})^T\boldsymbol{\Sigma}^{-1}(\mathbf{u}_{i}^{(g)}-\boldsymbol{\mu}) \\
    & = \sum^{n_g}_{i=1}[(\mathbf{u}_{i}^{(g)})^T\boldsymbol{\Sigma}^{-1}\mathbf{u}_i^{(g)}-(\mathbf{u}_{i}^{(g)})^T\boldsymbol{\Sigma}^{-1}\boldsymbol{\mu} \nonumber \\ 
    & \hspace{3cm}-\boldsymbol{\mu}^T\boldsymbol{\Sigma}^{-1}\mathbf{u}_i^{(g)} + \boldsymbol{\mu}^T\boldsymbol{\Sigma}^{-1}\boldsymbol{\mu}] \\
    & = \sum^{n_g}_{i=1}(\mathbf{u}_{i}^{(g)})^T\boldsymbol{\Sigma}^{-1}\mathbf{u}_i^{(g)}-\sum^{n_g}_{i=1}(\mathbf{u}_{i}^{(g)})^T\boldsymbol{\Sigma}^{-1}\boldsymbol{\mu} \nonumber \\ 
    & \hspace{2cm}-\sum^{n_g}_{i=1}\boldsymbol{\mu}^T\boldsymbol{\Sigma}^{-1}\mathbf{u}_i^{(g)} +n_g\boldsymbol{\mu}^T\boldsymbol{\Sigma}^{-1}\boldsymbol{\mu}
\end{align*}
Expanding $n_g\Delta^{(g)}$ then
\begin{align*}
    &n_g\Delta^{(g)}  = n_g(\boldsymbol{\mu}^{(g)}-\boldsymbol{\mu})^T\boldsymbol{\Sigma}^{-1}(\boldsymbol{\mu}^{(g)}-\boldsymbol{\mu}) \\
    & = n_g\left(\frac{1}{n_g}\sum^{n_g}_{i=1}\mathbf{u}_i^{(g)}-\boldsymbol{\mu}\right)^T\boldsymbol{\Sigma}^{-1}\left(\frac{1}{n_g}\sum^{n_g}_{i=1}\mathbf{u}_i^{(g)}-\boldsymbol{\mu}\right)\\
    & = \frac{1}{n_g}\left(\sum^{n_g}_{i=1}\mathbf{u}_i^{(g)}\right)^T\boldsymbol{\Sigma}^{-1}\left(\sum^{n_g}_{i=1}\mathbf{u}_i^{(g)}\right) - \sum^{n_g}_{i=1}(\mathbf{u}_{i}^{(g)})^T\boldsymbol{\Sigma}^{-1}\boldsymbol{\mu}\nonumber \\
    &\hspace{2.5cm}  -\sum^{n_g}_{i=1}\boldsymbol{\mu}^T\boldsymbol{\Sigma}^{-1}\mathbf{u}_i^{(g)}+n_g\boldsymbol{\mu}^T\boldsymbol{\Sigma}^{-1}\boldsymbol{\mu}
\end{align*}
This implies that
\begin{align*}
    &\delta^{(g)} - n_g\Delta^{(g)} \nonumber \\
    & = \sum^{n_g}_{i=1}(\mathbf{u}_{i}^{(g)})^T\boldsymbol{\Sigma}^{-1}\mathbf{u}_i^{(g)} - \frac{1}{n_g}\left(\sum^{n_g}_{i=1}\mathbf{u}_i^{(g)}\right)^T \boldsymbol{\Sigma}^{-1}\left(\sum^{n_g}_{i=1}\mathbf{u}_i^{(g)}\right) \\
    & = \frac{1}{n_g}\sum^{n_g}_{i=1}\mathbf{u}_{i}^{T}\boldsymbol{\Sigma}^{-1}(n_g\mathbf{u}_i-\sum^{n_g}_{\substack{j=1\\j\neq i}}\mathbf{u}_j) \\
    & = \frac{1}{n_g}\sum^{n_g}_{i=1}\mathbf{u}_{i}^{T}\boldsymbol{\Sigma}^{-1}\sum^{n_g}_{\substack{j=1\\j\neq i}}(\mathbf{u}_i-\mathbf{u}_j) \\
    & = \frac{1}{n_g}\sum^{n_g}_{1 \le i < j \le n_g}(\mathbf{u}_i^{(g)}-\mathbf{u}_{j}^{(g)})^T\boldsymbol{\Sigma}^{-1}(\mathbf{u}_i^{(g)}-\mathbf{u}_{j}^{(g)}) \\
    & = d^{(g)}_{\mathbf{c}}
\end{align*}
Therefore, 
\begin{equation*}
    \delta^{(g)}=d^{(g)}_{\mathbf{c}}+ n_g\Delta^{(g)},\quad g = 1,2,\dots,G.
\end{equation*}

\end{document}